\documentclass{article}
\usepackage[preprint]{main}
\usepackage[utf8]{inputenc} 
\usepackage[T1]{fontenc}    
\usepackage{hyperref}       
\usepackage{url}            
\usepackage{booktabs}       
\usepackage{amsfonts}       
\usepackage{nicefrac}       
\usepackage{microtype}      
\usepackage[utf8]{inputenc} 
\usepackage[T1]{fontenc}    
\usepackage{hyperref}       
\usepackage{url}            
\usepackage{booktabs}       
\usepackage{amsfonts}       
\usepackage{nicefrac}       
\usepackage{microtype}      
\usepackage{enumitem}
\usepackage{graphicx}
\usepackage{microtype}      
\usepackage{multirow}
\usepackage{graphicx}
\usepackage{caption}
\usepackage{pifont}
\usepackage{subcaption}
\usepackage{wrapfig}
\usepackage{authblk}
\usepackage{array}

\usepackage{colortbl}
\usepackage[table]{xcolor}

\usepackage{natbib}
\setcitestyle{numbers,square,comma}

\usepackage{float}
\usepackage{geometry}
\usepackage{placeins}

\title{AutoGeo: Automating Geometric Image Dataset Creation for Enhanced Geometry Understanding}

\author{
Zihan Huang$\thanks{Equal contribution.}$ \quad Tao Wu$^{*}$ \quad Wang Lin$^{*}$ \quad Shengyu Zhang \quad Jingyuan Chen$\thanks{Corresponding Author
.}$ \quad Fei Wu \\
Zhejiang University \\
\texttt{\{huanzh, twu22, linwanglw, sy\_zhang, jingyuanchen, wufei\}@zju.edu.cn}
}

\begin{document}

\maketitle

\begin{abstract}
With the rapid advancement of large language models, there has been a growing interest in their capabilities in mathematical reasoning. However, existing research has primarily focused on text-based algebra problems, neglecting the study of geometry due to the lack of high-quality geometric datasets. To address this gap, this paper introduces \textbf{AutoGeo}, a novel approach for automatically generating mathematical geometric images to fulfill the demand for large-scale and diverse geometric datasets. AutoGeo facilitates the creation of \textbf{AutoGeo-100k}, an extensive repository comprising $100$k high-quality geometry image-text pairs. By leveraging precisely defined geometric clauses, AutoGeo-100k contains a wide variety of geometric shapes, including lines, polygons, circles, and complex spatial relationships, etc. Furthermore, this paper demonstrates the efficacy of AutoGeo-100k in enhancing the performance of multimodal large language models through fine-tuning. Experimental results indicate significant improvements in the model's ability in handling geometric images, as evidenced by enhanced accuracy in tasks such as geometric captioning and mathematical reasoning. This research not only fills a critical gap in the availability of geometric datasets but also paves the way for the advancement of sophisticated AI-driven tools in education and research. Project page: \href{https://autogeo-official.github.io/}{https://autogeo-official.github.io/}.
\end{abstract}

\section{Introduction}
Mathematical reasoning is a critical component of human intelligence and a key objective of artificial intelligence (AI). The advancement of Multimodal Large Language Models (MLLMs), such as GPT-4~\cite{OpenAI_GPT4_2023} and LLaMa~\cite{touvron2023llama}, has demonstrated remarkable abilities in comprehension~\cite{zhan2024anygpt}, computation~\cite{yang2023gpt}, and reasoning~\cite{zhang2024multimodal}. Despite these advancements, the full utilization of MLLMs in mathematics, particularly in the area of geometric reasoning, has yet to be fully realized. 

Unlike algebra, which has been extensively studied~\cite{yuan2023large} and benefits from rich datasets~\cite{goswami2024aqua, hendrycks2021measuring}, geometry has received relatively little attention due to the lack of high-quality large-scale geometry datasets. Existing geometry datasets~\cite{geo3k, geos, geos++, GeoQA, GeoQA+, PGDP5k, PGPS9k} are primarily derived manually from examination papers or textbooks. As shown in Table~\ref{dataset compare}, these datasets constrain MLLMs' understanding of geometry due to their limited size and lack of detailed geometric descriptions. This limitation hinders the development of AI tools that can effectively understand geometric concepts and aid in personalized learning. Therefore, there is a clear need for the automatic creation of geometry datasets.

One potential approach to create the dataset is by utilizing diffusion models~\cite{ho2020denoising}. Although diffusion models have shown great potential for natural image synthesis and have addressed data shortage issues in many research areas~\cite{trabucco2023effective}, they face challenges in generating structured geometric images with coherent logical relationships (see Section~\ref{tbg}). Another approach involves using mathematical drawing software such as GeoGebra~\cite{10.1007/978-3-642-13166-0_95}, Matlab~\cite{MATLAB} and Matplotlib~\cite{Hunter:2007}. Attempts have been made to utilize large language models to automatically generate code for drawing geometries, but even advanced models like GPT-4 exhibit significant logical flaws, resulting in the production of distorted geometries as shown in Figure~\ref{fbg}. The generated prompts often fail to accurately describe the details of the geometric images. Furthermore, the process of collecting or writing a large number of diverse prompts is time-consuming and labor-intensive.

\begin{figure}
    \centering
    \includegraphics[width=\linewidth]{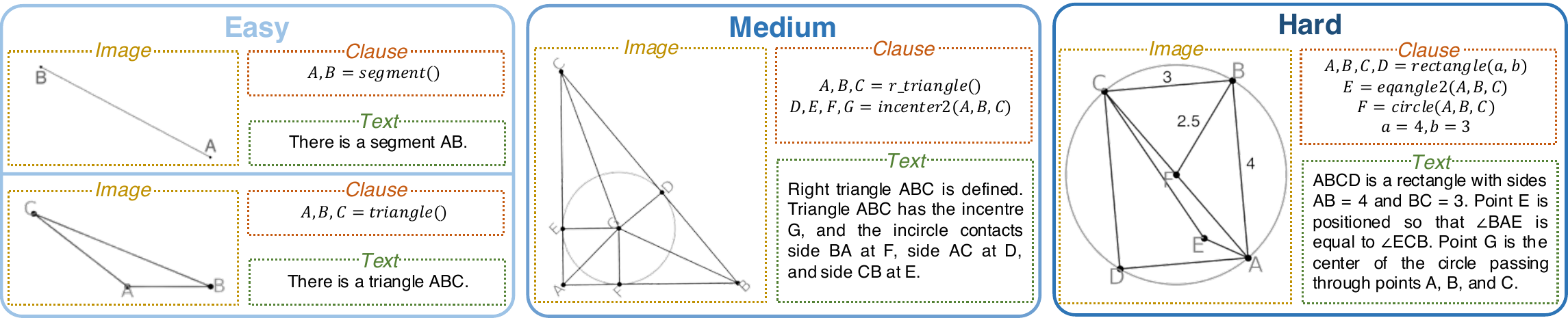}
    \vspace{-1.5em}
    \caption{Examples in AutoGeo-100k.}
    \label{example}
    \vspace{-1.5em}
\end{figure}

\begin{wraptable}{r}{0.5\textwidth}
  \vspace{-1em}
  \centering
  \caption{Comparison of AutoGeo-100k with mainstream geometry datasets.}
  \vspace{-0.5em}

    \scalebox{0.8}{
    \begin{tabular}{lccccc}
    \toprule
    \textbf{Datasets} & \textbf{\#Image-Text} & \textbf{Caption} & \textbf{Auto?}\\ 
    \midrule
    GEOS~\cite{geos} &  186  & \ding{55}  & \ding{55} \\ 
    GEOS++~\cite{geos++} & 1406  & \ding{55}  & \ding{55} \\ 
    Geometry3K~\cite{geo3k} & 3002 & \ding{55}  & \ding{55} \\ 
    GeoQA~\cite{GeoQA} & 4,998 & \ding{55}  & \ding{55} \\ 
    GeoQA+~\cite{GeoQA+} & 7,528 & \ding{55}  & \ding{55} \\ 
    UniGeo~\cite{UniGeo} & 9,543 & \ding{55}  & \ding{55} \\ 
    PGDP5k~\cite{PGDP5k} & 5,000 & \ding{55}  & \ding{55} \\ 
    PGPS9k~\cite{PGPS9k} & 9,022 & \ding{55}  & \ding{55} \\ 
    MathVerse~\cite{zhang2024mathverse} & 2,612 &  \ding{51}  & \ding{55} \\ 
    \midrule
    AutoGeo-100k & 100,000 &  \ding{51} & \ding{51} \\ 
    \bottomrule
    \end{tabular}%
    }
    \vspace{-1em}
  \label{dataset compare}%
\end{wraptable}

In this paper, we introduce \textbf{AutoGeo}, an innovative approach for automating the generation of geometric images, thereby facilitating the creation of extensive datasets at minimal expense. AutoGeo consists of three main components: the Augmented Geometry Clause System (AGCS), the Rule-based Clause Selector (RCS), and the Sample Generator (SG). The AGCS mitigates previous limitations by incorporating numerical clauses and categorizing them according to difficulty levels. The RCS selects compatible geometric clauses based on predefined rules to match desired complexities. The SG converts these clauses into data samples through two sub-modules: image generation utilizing Python to determine coordinates and apply transformations for diversity; and text generation employing ChatGPT to create descriptions for geometric images.

Based on AutoGeo, we construct a massive dataset named \textbf{AutoGeo-100k} which comprises 100k geometry image-caption pairs. As shown in Figure~\ref{example}, these images cover a broad spectrum of geometric structures while ensuring the integrity of geometric image data. We demonstrate the practical utility of our dataset by fine-tuning several MLLMs, resulting in improved abilities to comprehend geometric images. Through a series of experiments, we quantify the model's improved performance across various metrics, showcasing its enhanced accuracy in geometric caption and question-and-answering. 
In summary, our contributions are as follows:
\begin{itemize}[noitemsep,nolistsep,leftmargin=*]
\item
We propose \textbf{AutoGeo}, a novel system for the automated creation of geometric images and descriptions, addressing the longstanding challenge of dataset scarcity in the field of geometry.
\item
Leveraging AutoGeo, we efficiently construct a dataset of unprecedented scale, \textbf{AutoGeo-100k}, comprising 100k high-quality geometry image-text pairs while maintaining data integrity.
\item
We demonstrate the effectiveness of AutoGeo-100k through fine-tuning several MLLMs, significantly enhancing the models' capabilities in understanding geometric images.
\end{itemize}

\section{Related Work}
\subsection{Geometric Image Datasets}

Most existing datasets of geometry understanding and reasoning are constructed manually. They use image-text pairs in a Q\&A format to train and evaluate the geometric understanding and reasoning capabilities of multi-modal models. The GEOS~\cite{geos} dataset is one of the earliest efforts to systematize data in the field of geometry Q\&A, including $186$ plane geometry problems that encompass images, questions, and answers. The GEOS+~\cite{geos++} dataset expands on GEOS by increasing the number of geometry problems to $1,406$. Geometry3K~\cite{geo3k} collects more than $3,000$ SAT-style geometry problems from high school textbooks, covering a wider variety of geometries and problem types. GeoQA~\cite{GeoQA} and GeoQA+~\cite{GeoQA+} further expand the data volumes by adding annotations related to problem solving. PGDP5K~\cite{PGDP5k} contains more complex geometric elements and inter-element relationships. The largest geometric dataset is UniGeo~\cite{UniGeo}, which further expands the data size to $9$k and includes a more concise symbolic proof analysis process. PGPS9K~\cite{PGPS9k} includes more detailed diagram annotations and more solutions. The most recent dataset is Mathverse~\cite{zhang2024mathverse}, which contains $2,612$ problem samples and is labeled with detailed descriptions of image contents. 

The proposed AutoGeo framework overcomes the labor-intensive nature of previous efforts by introducing an automated, cost-effective pipeline for generating geometric images. This not only expands the current geometric image dataset but also addresses the issue of insufficient dataset size.

\subsection{Geometric Image Understanding and Reasoning}
Before the widespread adoption of MLLMs, earlier approaches~\cite{Inter-GPS, PGDP5k} have explored solutions for geometric problems. However, these approaches are limited by parameter constraints and lack robust reasoning abilities. For example, Inter-GPS~\cite{Inter-GPS} and PGDP~\cite{PGDP5k} employ symbolic methods, manually crafting geometric reasoning rules and symbol definitions for representing geometric objects. These models translate geometric images into symbols through techniques like instance segmentation~\cite{electronics12051199}, subsequently applying theorem search algorithms to derive solutions based on predefined rules. Recently, the advent of large language models has replaced manual theorem proving with powerful, data-driven reasoning. Projects such as GeoDRL~\cite{peng-etal-2023-geodrl}, G-Llava~\cite{gao2023gllava}, and SCA-GPS~\cite{ning2023symbolic} align geometric visual features with language model spaces, leveraging the inherent reasoning capabilities of these models instead of rule-based approaches.  Additionally, approaches like Alphageometry~\cite{trinh2024solving} combine both symbolic and language model reasoning for geometric theorem proving. 

Given the challenges in obtaining geometric image data, existing work based on MLLMs primarily focuses on enhancing the reasoning capabilities of language modules. Techniques such as chain-of-thought (CoT~\cite{zhang2024multimodal}) are utilized to improve the model's reasoning ability on geometry. Mathverse~\cite{zhang2024mathverse} reveals that current MLLMs still heavily rely on textual information for reasoning, with visual modules showing limited effectiveness. In contrast, our work focuses on enhancing model comprehension and reasoning for multimodal geometric problems by automating the synthesis of geometric images and descriptions, thereby improving the extraction and utilization of visual geometric information.

\section{AutoGeo: Automated Pipeline for Geometry Image Generation}
AutoGeo serves two essential needs for geometric image generation. First, it enables the generation of well-structured geometries by sampling from a predefined system of geometric clauses. Second, it allows for the cost-effective creation of a wide range of images with a high level of diversity. In this section we first present the performance of existing image generation baselines for generating geometries based on our trial-and-error experience. Then we describe AutoGeo's generation process. And finally, we show the static analysis of our dataset.
\begin{figure}
    \centering
    \includegraphics[width=\linewidth]{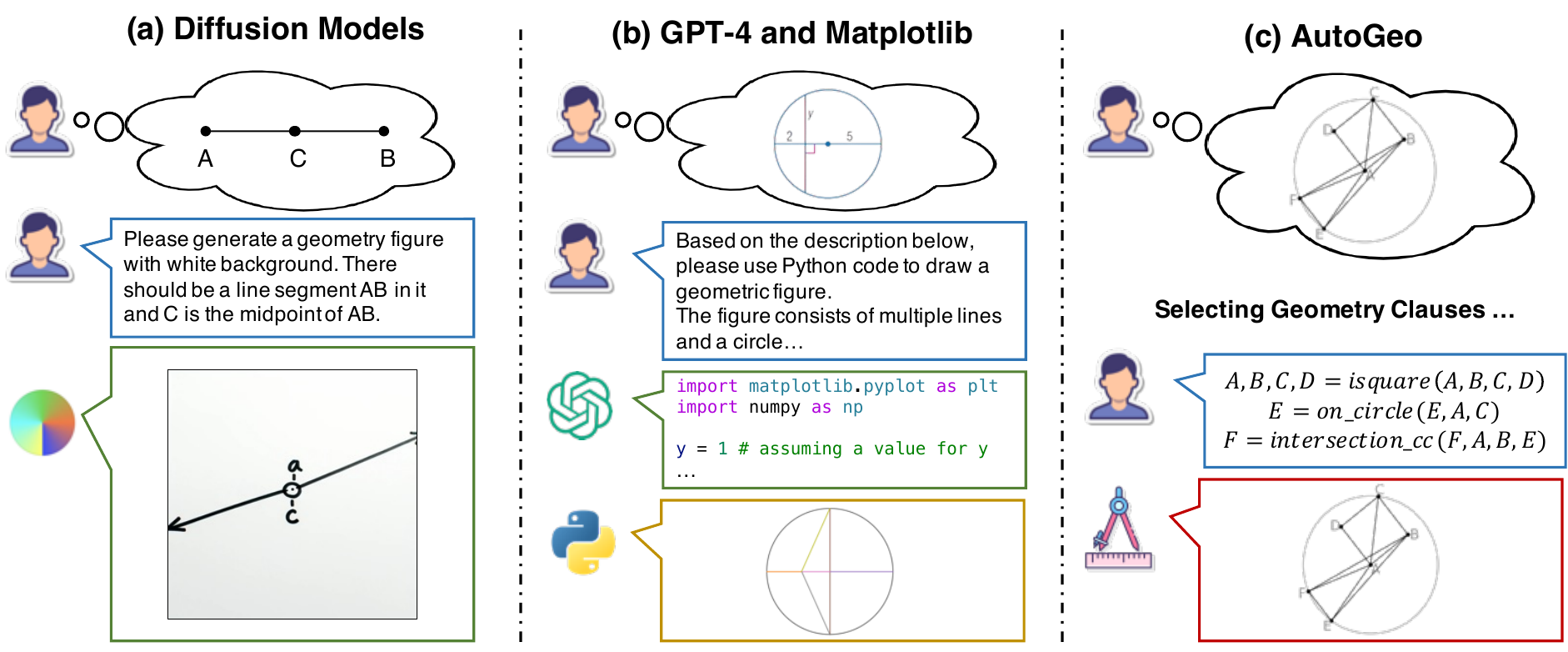}
    \vspace{-1.5em}
    \caption{We explore three approaches for generating geometric images. First, we utilize diffusion models, which encounter challenges in achieving precision due to the lack of a systematic logical generation process. Second, we employ GPT-4 to generate Python code alongside Matplotlib for image creation. However, this approach frequently encounters syntax and logical errors within the generated code. Consequently, we propose AutoGeo, an automatic geometry sample generation pipeline equipped with a comprehensive geometric clause system.}
    \label{fbg}
    \vspace{-1.5em}
\end{figure}

\subsection{Background: Diffusion Model or Python for Geometry Image Generation}
\label{tbg}
\textbf{Diffusion model.} 

Diffusion models~\cite{ho2020denoising} have emerged as a powerful approach in the field of image generation, offering a novel and effective method for creating high-quality synthetic images. 
The fundamental idea behind diffusion models is to model the distribution of data as a sequence of denoising steps. Starting from a noise vector, the model gradually refines the image by removing noise and enhancing details in each step.
However, when it comes to generating high-quality geometric images, diffusion models fall short. As illustrated in Figure \ref{fbg}(a), the model struggles to produce precise geometric images and even fails to draw straight lines. This limitation likely stems from the inherent differences between image types. Bitmap graphics, such as natural images, are composed of pixels arranged in a grid, making them suitable for the gradual refinement process of diffusion models. In contrast, vector graphics, like geometric images, are defined by mathematical equations, requiring a more logical and rigorous generation process that diffusion models currently cannot provide.

\textbf{Matplotlib.} Python Matplotlib package~\cite{Hunter:2007} is a tool specialized in creating static, animated, and interactive visualizations. With correct point coordination and Python code, Matplotlib can generate accurate geometric images. Manual coding, however, is labor-intensive, thus we resort to large language models to automate the code-writing process. As shown in Figure \ref{fbg}(b), we prompt GPT-4 with a natural language description of the image to generate the corresponding Python code. However, the resulting image often fails to match the original image and text description, indicating significant logical errors in the generated code. Additionally, basic syntax errors may exist in the code and further impede image generation. This reveals the current large language models' insufficient proficiency in generating code for geometric image generation.

Based on our explorations, we find that an effective automatic geometric image generation pipeline needs two key components: \textbf{1)} A comprehensive geometric definition system that includes basic geometric objects capable of constructing complex shapes, along with the formal geometric properties these points and lines must meet; and \textbf{2)} An accurate tool for drawing these geometric objects.

\subsection{AutoGeo Pipeline}
\begin{figure}
    \centering
    \includegraphics[width=\linewidth]{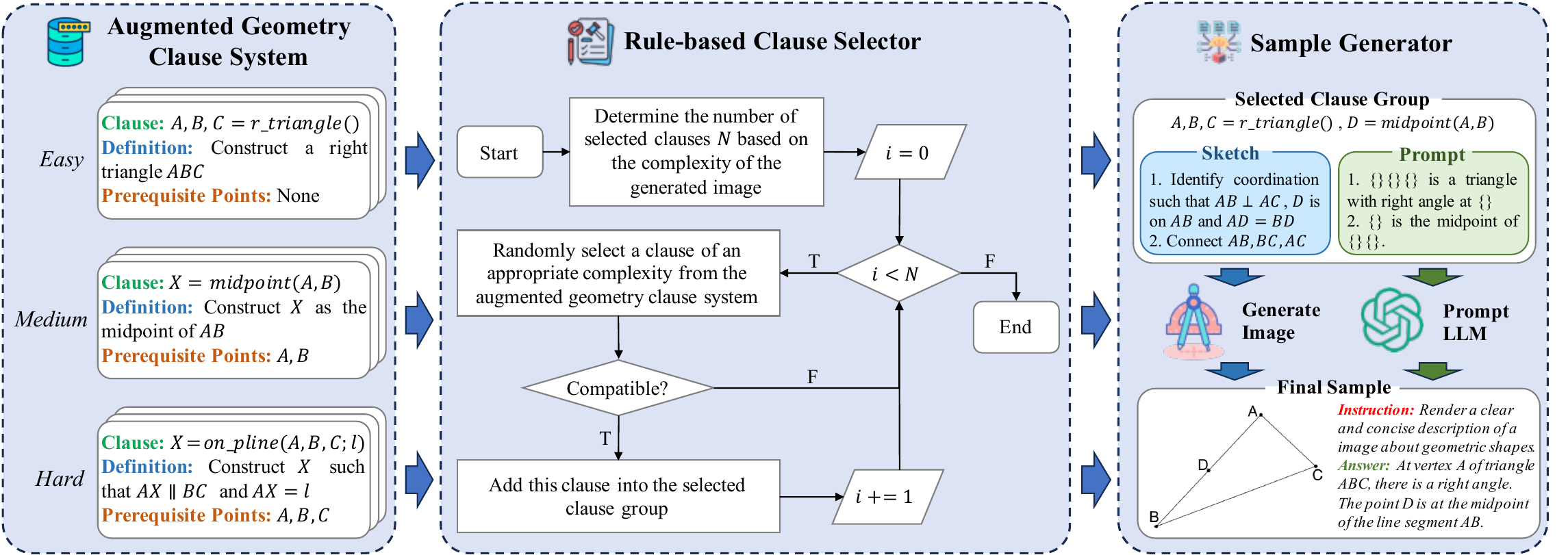}
    \vspace{-1.5em}
    \caption{Demonstration of AutoGeo pipeline. The augmented geometry clause system includes 77 clauses with key attributes. The system is enhanced by adding 26 clauses with numerical annotations and categorizing each clause into three levels of difficulty. The rule-based selector then automatically chooses mutually compatible clauses according to predefined rules to meet the complexity limits. Finally, the sample generator converts the selected clauses into dataset samples.}
    \label{fdataset}
    \vspace{-1.5em}
\end{figure}

In this section we introduce AutoGeo, a novel pipeline designed to automatically generate large-scale geometric datasets. As shown in Figure \ref{fdataset}, AutoGeo contains an augmented geometry clause system (Section~\ref{agcs}), a rule-based clause selector (Section~\ref{rcs}) and a sample generator (Section~\ref{sg}).

\subsubsection{Augmented Geometry Clause System}
\label{agcs}

\begin{figure}
    \centering
    \includegraphics[width=\linewidth]{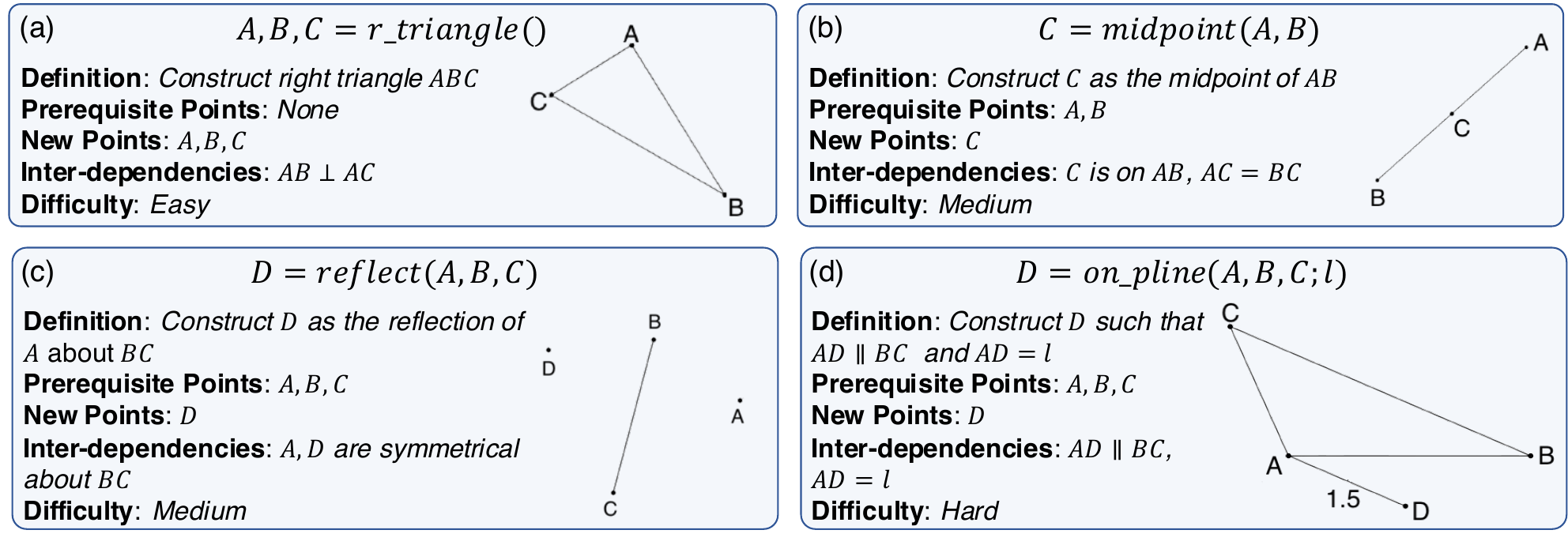}
    \vspace{-2em}
    \caption{Demonstrations of geometric clauses. Each geometric clause is a formalized description about a geometric definition, including (a) basic geometric objects, (b) properties of geometric objects, (c) geometric transforms and (d) geometric objects with numerical annotations. Each clause has several crucial attributes and has different difficulty levels, which facilitates the complexity control of the image generation process.}
    \label{fclause}
    \vspace{-1.5em}
\end{figure}

As introduced in Section~\ref{tbg}, an effective automatic geometric image generation pipeline should contain a comprehensive geometric definition system. Inspired by~\cite{trinh2024solving}, we reference its visualization system with geometric clauses as the foundation of our generation pipeline. A \textbf{geometric clause} is a formalized description of basic geometric objects, their properties or geometric transformations, which constitutes the fundamental units of complex geometric figures. As illustrated in Figure~\ref{fclause}, each geometric clause contains several vital attributes:

\begin{itemize}[noitemsep,nolistsep,leftmargin=*]
\item \textbf{Prerequisite points} are existing points necessary for constructing the geometric object. Please note that some geometric clauses do not require prerequisite points; we refer to these as ``\textit{Independent Clauses}'', like clause (a) in Figure \ref{fclause}. Conversely, clauses that do require prerequisite points are termed ``\textit{Dependent Clauses}'', such as clause (b) and (c) in Figure \ref{fclause}.
\item \textbf{Inter-dependencies} are geometric properties inherent in the geometric definitions, serving as essential conditions when generating geometric images based on the clauses. 
\end{itemize}

However, current geometry clause system still has some drawbacks. Firstly, it lacks geometry clauses that use numbers as input parameters. In geometric images, numerical annotations, such as segment lengths and angle sizes, are crucial for understanding the relationships between geometric objects and for further reasoning. Secondly, the system does not provide a clear complexity classification for each clause, which would aid in controlling the complexity of generated images.

To address these issues, we propose two enhancements to the current clause system. First, we investigate common geometric scenarios with numerical annotations and summarize 26 geometric clauses with numerical inputs. Training on geometric images with numerical annotations is expected to enhance the model's optical character recognition (OCR) capabilities and improve the reasoning performance. Second, we categorize the geometric clauses into three levels of difficulty: easy, medium, and hard. This classification will facilitate the complexity control of generated images in the rule-based clause selector. The final augmented geometry clause system comprises 77 clauses in total, categorized into 17 easy, 40 medium and 20 hard ones.

\subsubsection{Rule-based Clause Selector}
\label{rcs}
Based on the augmented geometry clause system, we introduce a rule-based clause selector. This selector uses predefined rules to automatically choose a series of mutually compatible geometric clauses that meet the target complexity.

\begin{wraptable}{r}{0.6\textwidth}
	\centering
 \vspace{-1.5em}
 \caption{Predefined rules for complexity control.}
 \vspace{-0.5em}
 \resizebox{0.6\textwidth}{!}{
	\begin{tabular}{c|c|ccc}
	\toprule
        \multirow{2}[2]{*}{Image Complexity} & \multirow{2}[2]{*}{Clause Number} & \multicolumn{3}{c}{Contains Clauses of Difficulty}\\
        \cmidrule{3-5}
        & & Easy & Medium & Hard \\
        \midrule\midrule
        Easy & $1$ &\checkmark & & \\
        Medium & $2$ & \checkmark & \checkmark&\checkmark (very few) \\
        Hard & $\geq 3$ & \checkmark & \checkmark&\checkmark \\
        \bottomrule
	\end{tabular}}
 \label{tcomplexity}
 \vspace{-1em}
\end{wraptable}

To create a geometric dataset with clear levels of difficulty, it is essential to control the number of geometric clauses in each sample. Thus initially, we determine the number of selected clauses $N$ based on the corresponding complexity. As the complexity of geometric images increases, the number of selected clauses grows accordingly. Next, we utilize the classification system in the augmented geometric clause system to facilitate the selection of clauses with appropriate difficulty. Generally, we avoid highly difficult clauses in geometric images of low complexity. Detailed rules of clause number and difficulty control are demonstrated in Table \ref{tcomplexity}.

Another important problem to consider is the compatibility of selected clauses, which depends primarily on their prerequisite points (defined in Section \ref{agcs}). For each chosen clause, we assess if there are sufficient points to meet its prerequisites. If the prerequisite points are adequate, the clause is deemed compatible with the currently selected clauses.

Once the rules for managing clause difficulty and their compatibility are established, the rule-based clause selector starts to automatically construct the selected clause group. It first determines the number of selected clauses based on the complexity of the generated image. Then it continuously selects clauses that match the required difficulty level and are compatible with the current system until reaching the set number.

\subsubsection{Sample Generator}
\label{sg}
The sample generator transforms selected clause groups into data samples. It consists of two sub-modules: image generation and text generation.

\textbf{Image generation.} We design a sketch function for each clause to efficiently convert it into geometric images. The sketch function is a piece of Python code, which first determines the coordinates of each new point defined in the clause. For \textit{dependent clauses} (defined in Section \ref{agcs}), the coordinates are easily derived from the prerequisite points and their inter-dependencies. For \textit{independent clauses}, we first create coordinates that satisfy the inter-dependency conditions. Then we apply geometric transformations, such as zooming and rotating, which preserve these inter-dependencies, to increase the diversity of the generated images.

Once the coordinates of each points are decided, the sketch function determines whether two points in the image should be connected, whether the connection should be a straight line or a curve, and whether it should be solid or dashed. This ensures the accuracy of the generated geometric images. Additionally, we apply two augmentations on the generated image to increase the task difficulty:

\begin{itemize}[noitemsep,nolistsep,leftmargin=*]
\item We assign different colors on each line and the background to enhance the diversity of the dataset;
\item We randomly mask small sections of the images as the absence of these small sections should not impact the model's comprehension of geometric images. Instead, the model should have the ability to reconstruct the missing parts based on the remaining geometric objects.
\end{itemize}

\textbf{Text generation.} We design 20 descriptive templates for each clause. For each clause in the selected group, we randomly choose and fill in a template, then request ChatGPT to combine and refine them. This process provides a diversified description for each image, serving as the ground truth response to the task instruction, ``\textit{Render a clear and concise description of an image about geometric shapes.}''

\subsection{AutoGeo-100k: Dataset Statistics and Characteristics}
Based on the proposed AutoGeo pipeline, we construct a large-scale geometric dataset, AutoGeo-100k. AutoGeo-100k contains 100k samples in total, encompassing 20k easy, 40k medium, and 40k hard ones. Figure~\ref{fsta} shows the frequency of each clause in data samples of different complexity levels and the average length of textual annotations corresponding to each clause. Detailed statistics are provided in Supplementary. AutoGeo-100k has three main characteristics:

\begin{figure}
    \centering
    \includegraphics[width=\linewidth]{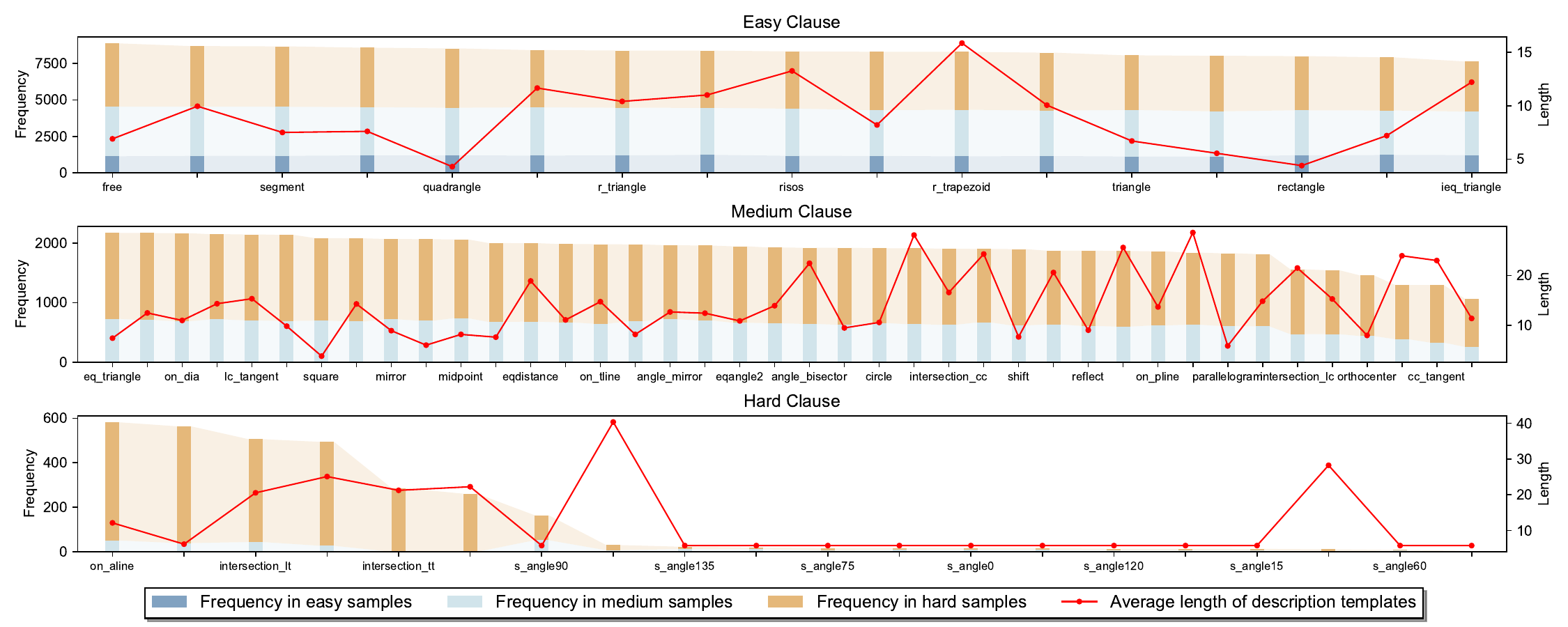}
    \vspace{-1em}
    \caption{Dataset statistics. The bar chart shows the frequency of each clause with different complexity levels. The line chart displays the average length of textual annotations corresponding to each clause. Half of the clause annotations are omitted in the figure for better visualization.}
    \label{fsta}
    \vspace{-1.5em}
\end{figure}

\begin{itemize}[noitemsep,nolistsep,leftmargin=*]
\item \textbf{Large-scale.} 
We use the AutoGeo pipeline to create a dataset of 100k geometric image-text pairs, surpassing the size of existing geometric datasets. Additionally, our clause-combination generation method theoretically allows for an unlimited number and complexity of samples.
\item \textbf{Low-cost.} AutoGeo is fully automatic and takes just 7.5 hours to generate a 100k-level dataset. This approach significantly reduces construction costs compared to human-annotation methods.
\item \textbf{Data validity.} AutoGeo pipeline employs a rigorously defined geometric clause system and an accurate visualization tool (\textit{i.e.}, Matplotlib), ensuring a precise one-to-one correspondence among clauses, geometric images and text annotations. This guarantees the validity of the data.
\end{itemize}

\section{Experiments}
We evaluate mainstream multimodal large language models (MLLMs) on AutoGeo-100k. Experiments on  geometric captioning and geometric question-and-answer (Q\&A) tasks demonstrate the limitations of existing MLLMs in geometric understanding and the effectiveness of AutoGeo-100k.

\subsection{Experiment Setup}

\textbf{Implementation details.}
For dataset generation, we utilize an Intel(R) Xeon(R) Gold 6240 CPU with 10 threads parallelism. In fine-tuning for captioning, we maintain the model's language module while adjusting the geometric semantic alignment through fine-tuning the model's projector layer and the LoRa layer~\cite{hu2021lora} in the vision module. To further finetune the model for geometric Q\&A task, we maintain the LoRa layer in the vision module and finetune the model's projector layer and Lora layer in language module on the augmented Geometry3K~\cite{gao2023gllava}. Our experiments are conducted on 8 A800s for 2 epochs, with a learning rate of 6e-5 and a batch size of 64.

\textbf{Baselines and metrics.} 
We conduct experiments on three MLLMs: LLaVA~\cite{llava}, InstructBLIP~\cite{dai2023instructblip}, and MiniGPT4-v2~\cite{chen2023minigptv2}. Additionally, we explore baseline models with varying sizes (LLaVA-7B and LLaVA-13B). For the geometry captioning task, we utilize Bleu~\cite{papineni-etal-2002-bleu}, ROUGE-L~\cite{lin-2004-rouge}, and CIDEr~\cite{vedantam2015cider} for evaluation. For the geometry Q\&A task, we utilize the average accuracy for evaluation.

\subsection{Experimental Results}
We experiment with fine-tuning various MLLMs on AutoGeo-100k to confirm that our dataset enhances geometric comprehension across models with diverse architectures and parameter sizes.

\begin{table}[]
\centering
    \setlength\tabcolsep{2pt}
    \begin{minipage}{0.56\textwidth}
    \centering
    \caption{Comparison of different MLLMs with zero-shot and fine-tuning (highlighted in grey) strategy in geometric captioning(AutoGeo-100k Test Set).}
    \scalebox{0.88}{
    \begin{tabular}{l|cccccc}
    \toprule
        Model & ROUGE-L & CIDEr & Bleu-1 & Bleu-2 & Bleu-3 & Bleu-4 \\
        \midrule\midrule
        LLaVA-7B & 11.41 & 0.18 & 9.77 & 2.77 & 0.86 & 0.35 \\
        \rowcolor[RGB]{230,230,230} LLaVA-7B & 24.68 & 51.96 & 28.68 &	16.55 & 11.30 & 8.34 \\
        LLaVA-13B & 10.04 & 0.10 & 8.11 & 2.05 & 0.44 & 0.14 \\
        \rowcolor[RGB]{230,230,230} LLaVA-13B & 23.50 & 45.50 & 26.30 & 15.27 & 10.43 & 7.61 \\
        MiniGPT4-v2 & 9.64 & 1.79 & 9.57 & 
1.75 & 0.39 & 0 \\
        \rowcolor[RGB]{230,230,230} MiniGPT4-v2 & 17.28 & 7.95 & 15.67 & 6.98 & 3.62 & 1.97 \\
        \bottomrule
    \end{tabular}
    }
    \label{tcaption}
\end{minipage}
    \hfill
    \begin{minipage}{0.4\textwidth}
    \centering
    \caption{Comparison in geometric Q\&A.} 

    \scalebox{0.83}{
    \begin{tabular}{l|c}
    \toprule
        Model & Accuracy \\
        \midrule\midrule
        Geoformer & 46.8  \\
        InstructBLIP & 49.2 \\ 
        UniMath & 50.0 \\
        \midrule
        LLaVA-7B & 18.40 \\
        \rowcolor[RGB]{230,230,230} LLaVA-7B+Geo3K & 49.73 \\
        \rowcolor[RGB]{230,230,230} LLaVA-7B+AutoGeo+Geo3K & 51.33\\ %
        LLaVA-13B & 22.50 \\
        \rowcolor[RGB]{230,230,230} LLaVA-13B+Geo3K & 52.79 \\
        \rowcolor[RGB]{230,230,230} LLaVA-13B+AutoGeo+Geo3K & 53.05 \\
        MiniGPT4-V2 & 21.30 \\
        \rowcolor[RGB]{230,230,230} MiniGPT4-V2+Geo3K & 27.98 \\
        \rowcolor[RGB]{230,230,230} MiniGPT4-V2+AutoGeo+Geo3k & 31.70
        \\
        \bottomrule
    \end{tabular}
    }
    \label{tsolving}
\end{minipage}
\end{table}

\begin{table}[]
    \centering
    \vspace{-1.5em}
    \caption{Experiments on training data difficulties.}
    \begin{tabular}{c|ccccccc}
    \toprule
        Data Volume & ROUGE-L & CIDEr & Bleu-1 & Bleu-2 & Bleu-3 & Bleu-4 & Accuracy \\
        \midrule\midrule
        LLaVA-v1.5-7B & 24.68 & 51.96 & 28.68 &	16.55 & 11.30 & 8.34 & 51.33 \\
        \midrule
        w/o easy data & 22.66 & 36.60 & 27.84 & 15.42 & 9.99 & 7.09 & 49.73 \\
        w/o middle data & 22.86 & 31.39 & 26.97 & 15.11 & 9.90 & 7.08 & 50.93 \\
        w/o hard data & 24.06 & 56.30 & 23.36 & 14.35 & 10.20 & 7.69 & 51.86 \\
        \bottomrule
    \end{tabular}
    \label{tdata}
    \vspace{-1.5em}
\end{table}
\textbf{Comparing different baselines on geometric captioning.}
Table~\ref{tcaption} presents the zero-shot and fine-tuning (highlighted in \textcolor{gray}{grey}) results of MLLMs on geometric captioning. As can be seen, prior to fine-tuning, these MLLMs generally struggle with captioning geometric images, particularly underperforming on the CIDEr metric. Specifically, the models tend to generate overgeneralized representations and produce verbose captions. LLaVA-7B exhibits better performance on the Bleu and ROUGE-L metrics, whereas the MiniGPT4-v2 model excells in the CIDEr metric. After fine-tuning, the models generate more concise and precise captions, exhibiting significant improvement across all captioning metrics.
This indicates the effectiveness of our AutoGeo-100k dataset in enhancing the models' ability to understand and describe geometric images.

\textbf{Comparing different baselines on geometric Q\&A.}
After fine-tuning MLLMs on AutoGeo-100k, we further fine-tune them using geometric Q\&A dataset, Geometry3K~\cite{gao2023gllava}. We evaluate the geometric Q\&A performance of these finetuned models on GeoQA~\cite{GeoQA} test set and compare their performance with task-specific models and general MLLM baselines. The results in the first part of Table~\ref{tsolving} indicate that general MLLMs' zero-shot performance on Q\&A task is lower than task-specific models. After fine-tuning on both captioning and QA data, MLLMs have notably improved performance on geometric QA and LLaVA-7B even surpasses the baselines specialized in geometric Q\&A.

\subsection{Ablation Study}
\begin{wrapfigure}{r}{0.45\textwidth}
        \centering
        \vspace{-1.5em}
        \includegraphics[width=0.4\textwidth]{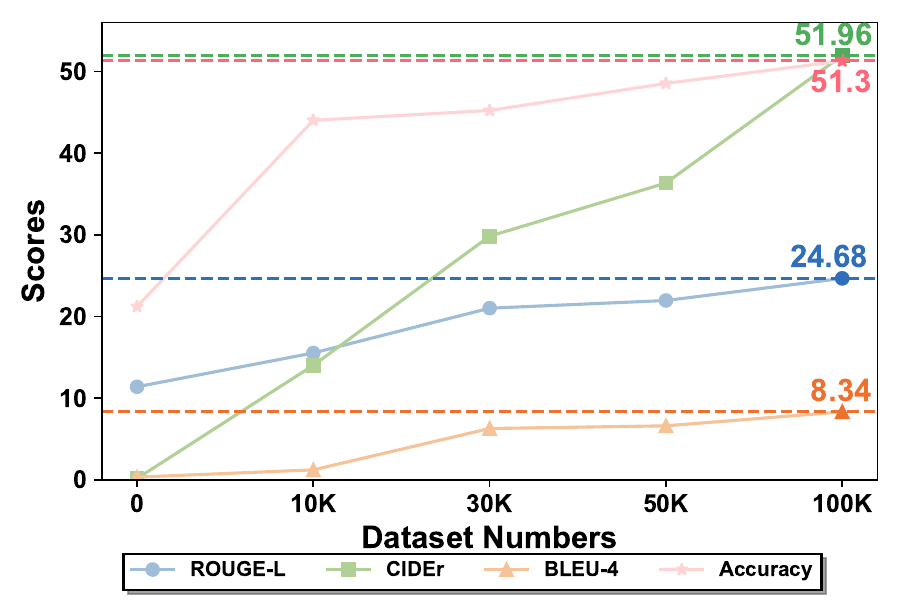}
        \vspace{-0.5em}
        \caption{Experiments on data volumes.}
        \vspace{-1.5em}
        \label{fe2}
\end{wrapfigure}
We conduct ablation studies on data volumes, data difficulties, and training components. 

\textbf{Ablation on data volumes.}
We conduct experiments to assess the impact of the training dataset volumes on model's performance. Specifically, we fine-tune LLaVA-7B with different  data volumes ranging from 10k to 100k. The result is in Figure~\ref{fe2}, demonstrating that model's performance on both geometric captioning and Q\&A improves continuously as the data volume grows. Even when the data volume reaches 100k, the performance still increases steadily. This trend suggests that even larger datasets will likely yield further performance improvements, revealing the necessity of AutoGeo's ability in efficiently generating large volumes of geometric training data. Detailed results are provided in Supplementary.

\textbf{Ablation on data difficulties.}
The AutoGeo-100k dataset includes 20k easy, 40k medium, and 40k hard samples. To assess the impact of each subset, we systematically remove one subset and replace it with an equal number of samples from the other difficulty levels, maintaining consistent training data volumes. The results, shown in Table \ref{tdata}, indicate that the absence of any difficulty level leads to a performance drop, highlighting the importance of diverse training data in complexity.

\begin{figure}
    \centering
    \includegraphics[width=\linewidth]{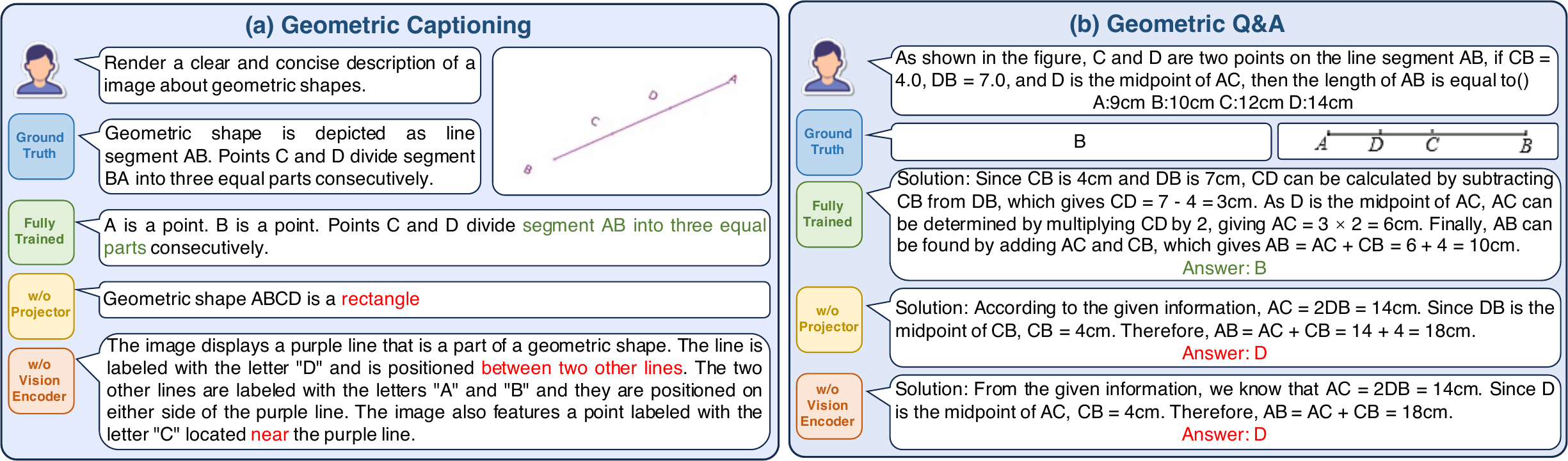}
    \caption{Visualization of ablation results on geometric captioning and geometric Q\&A tasks.}
    \label{fig_ablation}
    \vspace{-1.5em}
\end{figure}

\textbf{Ablation on fine-tuning strategies.}
We perform ablation studies to evaluate the impact of various fine-tuning strategies on geometric comprehension. The results in Table~\ref{tpipeline} demonstrate that fine-tuning both the vision and projection modules significantly enhances the model's performance on captioning tasks. 
Omitting fine-tuning the vision module (illustrated in Figure~\ref{fig_ablation}(a)) leads the model to overlook geometric concepts within the image, focusing instead on low-level visual pixels like the ``purple line''.
Omitting fine-tuning the projector module damages the model's descriptive capabilities, often resulting in imprecise and simplistic responses. As depicted in Figure~\ref{fig_ablation}(a), the model incorrectly describes the geometry containing the points ABCD as a rectangular ABCD. 
As shown in Figure~\ref{fig_ablation}(b), ablating either the vision module or the projector module diminishes the model's geometric Q\&A proficiency, leading to misunderstandings of the midpoint and generating incorrect answers.

\begin{table}[]
    \centering
    \vspace{-1em}
    \caption{Ablation studies on different fine-tuning strategies. +$\Delta$\textsubscript{projector} means fine-tuning on projector. +$\Delta$\textsubscript{vision encoder} means fine-tuning on vision encoder.}
    \scalebox{0.95}{
    \begin{tabular}{c|ccccccc}
    \toprule
        Data Volume & ROUGE-L & CIDEr & Bleu-1 & Bleu-2 & Bleu-3 & Bleu-4 & Accuracy \\
        \midrule\midrule
        LLaVA-v1.5-7B & 11.41 & 0.18 & 9.77 & 2.77 & 0.86 & 0.35 & 21.22\\
        +$\Delta$\textsubscript{projector} & 22.34 & 30.13 & 27.06 & 14.61 & 8.94 & 5.69 & 50.53 \\ 
        +$\Delta$\textsubscript{vision encoder} & 23.22 & 43.57 & 27.52 & 15.35 & 10.29 & 7.50 & 46.29 \\
        +$\Delta$\textsubscript{vision encoder}+$\Delta$\textsubscript{projector} & 24.68 & 51.96 & 28.68 & 16.55 & 11.30 & 8.34 & 51.33\\    
        \bottomrule
    \end{tabular}
    }
    \label{tpipeline}
\end{table}

\section{Conclusion}
\textbf{AutoGeo} is featured as an automated pipeline that efficiently generates a diverse and large-scale dataset with minimal cost. This approach addresses the previously limited availability of high-quality geometric datasets and provides a foundation for research and application in AI-driven educational tools and beyond.
Based on AutoGeo, we construct \textbf{AutoGeo-100k} with a collection of 100k high-quality geometry image-text pairs. AutoGeo-100k provides massive data for training and evaluating multimodal large language models (MLLMs). Our experiments demonstrate that fine-tuning MLLMs on AutoGeo-100k substantially enhances their performance on tasks such as geometric captioning and question-and-answering, highlighting the dataset's ability to improve models' understanding of geometric concepts.
Looking forward, the AutoGeo pipeline and the AutoGeo-100k dataset will facilitate further exploration and development in geometric reasoning. We hope that the dataset will inspire innovations in multimodal understanding and contribute to the advancement of AI systems capable of mathematical reasoning.

\appendix

\counterwithin{figure}{section}
\counterwithin{table}{section}

\renewcommand{\thefigure}{\arabic{figure}}
\renewcommand{\thetable}{\arabic{table}}

\clearpage
\newpage

\bibliography{main}
\bibliographystyle{unsrt}
\end{document}